\documentclass[letterpaper]{article} 
\usepackage{aaai25}  
\usepackage{times}  
\usepackage{helvet}  
\usepackage{courier}  
\usepackage[hyphens]{url}  
\usepackage{graphicx} 
\usepackage{natbib}  
\usepackage{caption} 
\usepackage{newfloat}
\usepackage{listings}
\DeclareCaptionStyle{ruled}{labelfont=normalfont,labelsep=colon,strut=off} 
\usepackage{cancel}
\usepackage{amssymb}
\usepackage{amsmath}
\usepackage{booktabs}
\usepackage{subcaption}
\usepackage{xcolor}
\usepackage{tkz-graph}
\usepackage{pgfplots}
\pgfplotsset{compat=newest}
\usepackage[procnumbered,ruled,linesnumbered,vlined ]{algorithm2e} 
\SetKwInput{KwInput}{Inputs}
\SetKwInput{KwOutput}{Output}
\SetKwComment{Comment}{$\triangleright$\ }{}
\newcounter{algoline}

\SetNlSty{bfseries}{\color{black}}{}
\makeatletter
\newcommand{\algorithmfootnote}[2][\footnotesize]{%
  \let\old@algocf@finish\@algocf@finish
  \def\@algocf@finish{\old@algocf@finish
    \leavevmode\rlap{\begin{minipage}{\linewidth}
    #1#2
    \end{minipage}}%
  }%
}
\SetKwFor{ParallelFor}{for}{do in parallel}{end for}
\SetKwFor{ForEach}{for each}{do}{end}
\title{Parallelizing Multi-objective {A*} Search}
\author{Saman Ahmadi\textsuperscript{*\rm 1},
Nathan R. Sturtevant\textsuperscript{\rm 2},
Andrea Raith\textsuperscript{\rm 3},
 Daniel Harabor\textsuperscript{\rm 4},
Mahdi Jalili\textsuperscript{\rm 1}}
\affiliations{
\textsuperscript{\rm 1} School of Engineering, RMIT University, Australia\\
\textsuperscript{\rm 2} Department of Computing Science, University of Alberta\\
\textsuperscript{\rm 3} Department of Engineering Science, University of Auckland, New Zealand\\
\textsuperscript{\rm 4} Department of Data Science and AI, Monash University, Australia\\
    *saman.ahmadi@rmit.edu.au
}

\usepackage{bibentry}

\begin{document}

\maketitle

\begin{abstract}
The Multi-objective Shortest Path (MOSP) problem is a classic network optimization problem that aims to find all Pareto-optimal paths between two points in a graph with multiple edge costs. 
Recent studies on multi-objective search with A* (MOA*) have demonstrated superior performance in solving difficult MOSP instances.
This paper presents a novel search framework that allows efficient parallelization of MOA* with different objective orders.
The framework incorporates a unique upper bounding strategy that helps the search reduce the problem's dimensionality to one in certain cases.
Experimental results demonstrate that the proposed framework can enhance the performance of recent A*-based solutions, with the speed-up proportional to the problem dimension.
\end{abstract}

\section{Introduction}
The Multi-objective Shortest Path problem (MOSP), a challenging network optimization problem, aims to 
find all Pareto-optimal paths between a given pair of nodes in a network where each edge is associated with multiple attributes.
MOSP can emerge in various real-world problems.
Applications are multi-objective path planing for mobile robots \cite{RenRC21}, multi-objective route selection problem for unmanned air vehicles \cite{TezcanerK11}, and multi-objective routing for airport ground movement with factors such as taxi time, fuel consumption and emissions \cite{weiszer2020multi}.

\citet{SalzmanF0ZCK23} presented an overview of recent advances in bi-objective and multi-objective search, highlighting the significant progress made by heuristic search in enhancing the efficiency of \textsf{MOA*} \cite{StewartW91}.
The \textsf{EMOA*} \cite{EMOA22}, \textsf{TMDA} \cite{TMDA23} and \textsf{LTMOA*} \cite{LTMOA3} algorithms are three state-of-the-art approaches that utilize best-first search to solve point-to-point MOSP more efficiently.
The \textsf{LTMOA*} algorithm, in particular, is shown to perform up to an order of magnitude faster than \textsf{EMOA*} due to its more efficient dominance checking rules.
Nevertheless, \textsf{LTMOA*}'s performance in large networks degrades when the number of objectives increases, leaving some difficult MOSP instances unsolved even after a one-hour runtime.
Recently, the \textsf{NWMOA*} algorithm \cite{AhmadiSHJ24} has demonstrated faster performance than previous approaches, including the improved variant of the \textsf{NAMOA*\textsubscript{dr}} algorithm \cite{NAMOA15} studied in \citet{TMDA23}.

In many domains, performance of planning algorithms may be impacted by slight modifications of the search setting, such as changes on tie-breaking rules or order of operators \cite{Knight93,HoweD02,AhmadiTHK23_Networks}. 
In the case of multi-objective search, while \textsf{MOA*} offers a robust framework for optimally solving MOSP in large graphs, the order of objectives can have a ``\textit{dramatic effect on the algorithm’s running times}" \cite{SalzmanF0ZCK23}. 
Although a good-performing ordering can sometimes be obtained empirically, as in \citet{LTMOA3}, there is no guarantee that such ordering performs best in all instances. 
Incorporating multiple objective orderings in a parallel setting can be seen as a potential solution to the above shortcoming.
Despite being well-studied for single-objective search in various settings \cite{ValenzanoSSBK10,ZhouZ15,FukunagaBJK18}, parallelization remains largely underexplored in the literature on multi-objective search.
While existing attempts mainly focus on parallelization of search procedures \cite{SandersM13,ErbKS14,Ulloa0KFS24}, the bi-objective search algorithm \textsf{BOBA*} \citet{AhmadiTHK21_esa} offers a parallel framework that executes two bi-objective searches on different objective orderings.
This approach has shown to accelerate the standard bi-objective A* search, however, its applicability for parallelizing \textsf{MOA*} remains uncertain. 

This research proposes a parallel search framework for \textsf{MOA*}, representing, to the best of our knowledge, the first attempt to parallelize multi-objective search for more than two objectives.
Inspired by the search scheme of \textsf{BOBA*}, our proposed approach runs multiple \textsf{MOA*} searches in parallel, but on different objective orderings.
Central to our framework is an innovative upper bounding technique that not only shrinks the scaled search space but also reduces the problem's dimensionality to one under specific conditions.
Results demonstrate the effectiveness of our parallel framework in achieving scaled acceleration of the \textsf{MOA*} search, with speed-ups proportional to the number of parallel searches, achieving up to an order of magnitude improvement on challenging MOSP instances.

\section{Notation and Problem Formulation}
Consider a directed graph $G=(S,E)$ with a finite set of states $S$ and a set of edges $E \subseteq S \times S$ with every edge $e \in E$ comprising $\mathit{k} \in \mathbb{N}$ attributes that can be accessed via the cost function $\mathbf{cost}:E \to \mathbb{R}^k$, that is, we have ${\mathbf{cost}}=(\mathit{cost}_1, \mathit{cost_2}, \dots, \mathit{cost_k})$ as a form of vector.
A path $\pi$ is a sequence of states $u_i \in S$ with $ i \in \{1, \dots, n \}$ and $(u_i,u_{i+1}) \in E | \ i < n$.
The $\mathbf{cost}$ vector of the path is then the sum of corresponding attributes on all the edges constituting the path, namely $\mathbf{cost}(\pi) = \sum_{i=1}^{n-1}{\mathbf{cost}(u_i,u_{i+1})}$.
The MOSP problem aims to find a set of Pareto-optimal paths between a given pair of $\mathit{start}\in S$ and $\mathit{goal}\in S$. 
A path $\pi^*$ is Pareto efficient if there does not exist any other path $\pi'$ 
that simultaneously satisfies $\mathit{cost}_i(\pi')<\mathit{cost}_i(\pi^*)$ and $\mathit{cost}_j(\pi')\leq\mathit{cost}_j(\pi^*)$ for any $i,j \in \{1,\dots,k\} | i\neq j$.

In the context of A* search, we define our search objects as \textit{nodes}.
A node $x$ is a tuple that contains the main information on the partial path from $\mathit{start}$ to state $s(x)$, where $s(x)$ is a function that returns the state associated with $x$. 
Node $x$ contains: i) the cost vector $\mathbf{g}(x)$, representing the $\mathbf{cost}$ of a concrete path from the $\mathit{start}$ state to state $s(x)$; ii) the cost vector $\mathbf{f}(x)$, representing the $\mathbf{cost}$ estimate of a complete path from $\mathit{start}$ to $\mathit{goal}$ via $s(x)$; iii) the reference node $\mathit{parent}(x)$, indicating the parent node of $x$.

All operations of the cost vectors are considered to be performed element-wise.
For example, we define ${\bf g}(x) + {\bf g}(y)$ as $\left({g_1}(x) + {g_1}(y), \dots, {g_{k}}(x) + {g_{k}}(y)\right)$.
We use $\preceq$ or $\prec$ symbols in direct comparisons of cost vectors, e.g. ${\bf f}(x) \preceq {\bf f}(y)$ denotes ${f_i}(x) \leq {f_i}(y)$ for all $i \in \{1,\dots,k\}$.
Further, the operator $\mathrm{Tr}(\mathbf{v})$ truncates the first cost of the vector $\mathbf{v}$, and $\mathrm{Tr}^\lambda$ denotes $\lambda$ consecutive truncations, e.g., $\mathrm{Tr}({\bf f}(x))=\left({f_2}(x), \dots, {f_{k}}(x)\right)$ and $\mathrm{Tr}^2({\bf f}(x))=\left({f_3}(x), \dots, {f_{k}}(x)\right)$.

\noindent \textbf{Definition\ }
Cost vector $\mathbf{v}$ is weakly dominated by cost vector $\mathbf{v}'$ if we have $\mathbf{v}'\preceq\mathbf{v}$; $\mathbf{v}$ is dominated by $\mathbf{v}'$ if $\mathbf{v}' \preceq \mathbf{v}$ and $\mathbf{v}' \neq \mathbf{v}$; $\mathbf{v}$ is not dominated by $\mathbf{v}'$ if $\mathbf{v}' \npreceq \mathbf{v}$.
Node $y$ is weakly dominated by node $x$ if ${\bf f}(x) \preceq {\bf f}(y)$.

\section{Multi-objective Search with A*}
Multi-objective pathfinding with A* involves a systematic search by \textit{expanding} nodes in best-first order, that is, the search is guided by paths showing smallest $\mathit{start}$-$\mathit{goal}$ cost estimates.
These estimates are traditionally established as ${\bf f}(x)={\bf g}(x)+{\bf h}(s(x))$ where a consistent heuristic function ${\bf h}: S \rightarrow \mathbb{R}^k$ estimates lower bounds on the $\mathbf{cost}$ of extended paths to $\mathit{goal}$ \cite{hart1968formal}.

To elaborate on the key steps of the search, we have provided in Algorithm~\ref{alg:moa} a pseudocode of the recent \textsf{MOA*} methods to MOSP, namely \textsf{LTMOA*} and \textsf{NWMOA*}.
The search starts with initializing a pair of node sets.
The first set, $\mathit{Open}$, is responsible for maintaining unexplored nodes in best-first order.
The second set, $\mathit{Sols}$, stores all discovered solution nodes during the search, and is returned as output.
Next, it initializes, for every state of the graph, a list called $\mathrm{G^{Tr}}$.
For a typical state $u$, $\mathrm{G^{Tr}}(u)$ always contains non-dominated truncated cost vectors expanded with $u$. 
To begin the search, the algorithm then initializes a node with the $\mathit{start}$ state and inserts it into $\mathit{Open}$.  

\begin{algorithm}[!t]

\small
\caption{Multi-objective Search with {A*} }

\label{alg:moa}
\DontPrintSemicolon
 \KwIn{A MOSP problem ($\mathit{G}$, $\mathit{\bf cost}$, ${\bf h}$, $\mathit{\mathit{start}}$, $\mathit{\mathit{goal}}$)}
 
 \KwOutput{A cost-unique Pareto-optimal solution set}
 
 $\mathit{Open} \gets \emptyset $ , $\mathit{Sols} \gets \emptyset $\;
$ \mathrm{G^{Tr}}(u) \gets \emptyset$ \ $\forall u \in S$\ \;

 $x \gets $ new node with $s(x) = \mathit{start}$\ \;
 $ {\bf g}(x) \gets {\bf 0}$ ,
 $ {\bf f}(x) \gets {\bf h}(\mathit{start})$ ,
 $parent(x) \gets null $ \;
Add $x$ to $Open$\;

\While{$Open \neq \emptyset$ \label{alg:moa:iter}}
{
 
Extract node $x$ from $Open$ with the smallest $\textbf{f}$-value \label{alg:moa:least_cost} \;

       \If{ $\mathtt{IsDominated}(\mathrm{Tr}({\bf g}(x)),\mathrm{G^{Tr}}(s(x)))$ \textnormal{\bf or}  $\mathtt{IsDominated}(\mathrm{Tr}({\bf f}(x)),\mathrm{G^{Tr}}(\mathit{goal}))$ \label{alg:moa:dom}} 
      {\textbf{continue} }
      
     $\mathtt{RemoveDominated}(\mathrm{Tr}({\bf g}(x)), \mathrm{G^{Tr}}(s(x)))$ \label{alg:moa:remove_dom}\;

    Add $\mathrm{Tr}({\bf g}(x))$ to $\mathrm{G^{Tr}}(s(x))$ \label{alg:moa:add_to_list}\;
     
\If{$ s(x) = \mathit{goal} $ \label{alg:moa:goal} \label{alg:moa:sol0}}
    {      
        Add $x$ to $Sols$ \label{alg:moa:sol3}\;
        {\bf continue} \;
    }
    
    \ForEach{$t \in Succ(s(x))$}
        {  $y \gets $ new node with $s(y) = t$ \label{alg:moa:expansion1}\; 
             ${\bf g}(y) \gets {\bf g}(x) + {\bf cost} (s(x),t)$ \; 
             ${\bf f}(y) \gets {\bf g}(y) + {\bf h} (t)$ \;
             $parent(y) \gets x$ \label{alg:moa:expansion2}\; 
        \If{ $\mathtt{IsDominated}(\mathrm{Tr}({\bf g}(y)),\mathrm{G^{Tr}}(t))$ \textnormal{\bf or}  $\mathtt{IsDominated}(\mathrm{Tr}({\bf f}(y)),\mathrm{G^{Tr}}(\mathit{goal}))$ \label{alg:moa:dom2}} 
      {\textbf{continue} }
        Add $y$ to $Open$ \label{alg:moa:expansion3}\;
        }
}
\Return{$Sols$}
\end{algorithm}

Each iteration of the algorithm starts at line~\ref{alg:moa:iter}.
Let $x$ be a node offering the smallest ${\bf f}$-value among nodes in $\mathit{Open}$.
Within each iteration, the algorithm attempts to extend $x$ towards $\mathit{goal}$.
However, to avoid processing unpromising nodes, $x$ is checked for dominance through the $\mathtt{IsDominated}$ procedure (line~\ref{alg:moa:dom}).
The dominance rules are: i) whether $x$ is weakly dominated by any node previous expanded with $s(x)$; ii) whether $x$ can be weakly dominated by any current solution.
Weakly dominated nodes can be safely pruned, as their expansion would not lead to any cost-unique optimal solution.
Since nodes in \textsf{MOA*} are explored in non-decreasing order of their primary cost estimate (here $f_1$), dominance checks can be done more efficiently by comparing the non-primary costs only.
Algorithm~\ref{alg:moa} utilizes this dimension reduction technique within its dominance rules by truncating the first cost of the vector using the $\mathrm{Tr}(\cdot)$ operator.
Given $\mathrm{G^{Tr}}(s(x))$ containing the (non-dominated) truncated cost vectors of previous nodes explored with $s(x)$, node $x$ is pruned if $\mathrm{Tr}({\bf g}(x))$ is weakly dominated by a cost vector in $\mathrm{G^{Tr}}(s(x))$.
Similarly, $x$ is pruned if its truncated $\mathbf{cost}$ estimate $\mathrm{Tr}({\bf f}(x))$ is weakly dominated by that of any solution in $\mathrm{G^{Tr}}(\mathit{goal})$.
Otherwise, if $x$ is not pruned by either of the rules, its truncated $\mathbf{cost}$ vector will be stored for the purpose of future dominance tests with $s(x)$ (line~\ref{alg:moa:add_to_list}),
but before that, the algorithm removes from $\mathrm{G^{Tr}}(s(x))$ all vectors dominated by $\mathrm{Tr}({\bf g}(x))$.
This operation ensures the vectors of $\mathrm{G^{Tr}}(s(x))$ always remain non-dominated.

Once proven non-dominated, $x$ will be either added to $\mathit{Sols}$ if $s(x)$ is the $\mathit{goal}$ state (line~\ref{alg:moa:sol0}), or expanded.
The expansion operation \textit{generates} a set of successor states, each denoted $\mathit{Succ}(s(x))$.
Let $y$ be a descendant node of $x$. 
To reduce the queue load, \textsf{MOA*} can check $y$ against dominance rules before inserting it into $\mathit{Open}$ for further expansions (line~\ref{alg:moa:dom2}).
However, the literature has documented faster \textsf{MOA*} performance with \textit{lazy} dominance checks, where nodes are checked against dominance rules only after they are extracted from $\mathit{Open}$.
Finally, \textsf{MOA*} terminates when there is no node in $\mathit{Open}$ to explore, returning $\mathit{Sols}$ as a set of cost-unique solution nodes to the given MOSP instance.

\textbf{\textsf{LTMOA*} vs. \textsf{NWMOA*}:}
Both algorithms follow the outline of \textsf{MOA*} in Algorithm~\ref{alg:moa}, but they differ in their node exploration methods in three aspects:\\
i) \textbf{Ordering of nodes:}
The priority queue in \textsf{LTMOA*} orders nodes based on their $\mathbf{cost}$ lexicographically, whereas \textsf{NWMOA*} explores nodes in the order of primary cost only.
The former reduces node expansions, with no expansions of dominated nodes.
The latter, however, may incur extra expansions but does not need to handle tie-breaking.\\
ii) \textbf{Dominance check:}
\textsf{LTMOA*} stores the truncated cost vector of previous expansions in no specific order, whereas \textsf{NWMOA*} stores them in lexicographical order.
The former necessitates a linear scan over all vectors of the $\mathrm{G^{Tr}}$ list to ensure the new vector is non-dominated but allows non-dominated vectors to be added to the list more efficiently.
The latter, however, allows for partial traversal over vectors of the  $\mathrm{G^{Tr}}$ list expansions, but incurs sorting overhead.\\
iii) \textbf{Quick pruning:}
\textsf{NWMOA*} utilizes an additional dominance test that enables quick pruning of some nodes.
At every state it caches the most recent non-dominated truncated cost vector, which is used as a first (more-informed) candidate for quick dominance pruning before attempting the full dominance tests.
This technique has proven to be effective in reducing the total number of dominance checks drastically, enhancing the search performance.
\section{A Parallel \textsf{MOA*} Framework}
This section describes our novel parallelized \textsf{MOA*} solution to MOSP.
The idea is straightforward: let \textsf{MOA*} be guided with more than one objective order at a time.
Here, we are interested in a set of orderings that allows every objective to appear as the primary cost in one search. 
Algorithm~\ref{alg:high-level} presents the higher level of the proposed parallel \textsf{MOA*}.
Given a $k$-dimensional MOSP instance, the algorithm uses $k$ CPU threads to run $k$ individual \textsf{MOA*} searches in parallel such that the $i$-th thread
, $i \in \{1,\dots,k\}$, 
is provided with a cost and heuristic function that return $\mathit{cost}_i$ and $h_i$ as primary cost and heuristic, respectively (lines~\ref{alg:high-level:mult-search1}-\ref{alg:high-level:mult-search3}).
One such set of orderings can simply be all cyclic permutations of the costs.
For example, in a three-dimensional instance ($k= 3$), the second thread can use a cost function that returns edge $\mathbf{cost}$ in $(\mathit{cost}_2, \mathit{cost_3}, \mathit{cost_1})$ order with $\mathit{cost}_2$ as the primary cost.
For this thread, we provide \textsf{MOA*}'s heuristic function ${\bf h}$ in $(h_2,h_3,h_1)$ order.
Since each \textsf{MOA*} search is complete, the parallel loop can terminate as soon as one thread finishes its search.

\textbf{Necessity of upper bounding:}
Although the simple parallelization above allows for more than one objective ordering to be involved, it does not necessarily reduce the overall computation time if the searches are conducted independently, primarily due to algorithmic overhead driven by multiple computationally demanding \textsf{MOA*} searches.
More precisely, upon the termination of the search in one thread, we have captured all optimal solutions, rendering the search effort of other threads superfluous.
To reduce this overhead, 
the search in each thread could be informed with the optimal solutions discovered in the other threads, so they can prune unpromising paths not leading to a $\mathit{start}$-$\mathit{goal}$ path better than \textit{any} discovered optimal solution as global upper bounds.
Nonetheless, this method is not efficient in practice, essentially because upper bounding via all discovered solutions (in linear-time fashion) becomes costly in the absence of a unified objective ordering.
Consequently, effective upper bounding remains a crucial bottleneck for efficient parallelization of \textsf{MOA*}.

\begin{algorithm}[t]
\small
\caption{Parallel \textsf{MOA*}- Higher Level}
\label{alg:high-level}
\DontPrintSemicolon
\KwIn{A MOSP Problem ($\mathit{G}$, $\mathit{\mathit{start}}$, $\mathit{\mathit{goal}}, k$)}

\KwOutput{A cost-unique Pareto-optimal solution set}

$\mathit{\bf Sols} \gets \emptyset^k$, $\overline{\bf f} \gets \infty^k$\label{alg:high-level:init}\;
    \ParallelFor{$ i \in \{1, \dots, k \}$}
    {  
        $\mathit{\bf c} \gets$ cost function with $\mathit{cost}_i$ as primary cost\label{alg:high-level:mult-search1}\;
        
        $ {\bf h} \gets$ heuristic function corresponding to $\mathit{\bf c}$\label{alg:high-level:mult-search2}\;
        
        Parallelized \textsf{MOA*}\ on ($\mathit{G}$, $\mathit{\bf c}$, ${\bf h}$, $\mathit{\mathit{start}}$, $\mathit{\mathit{goal}}$, $\mathit{Sols}_i$)
        \label{alg:high-level:mult-search3}
    }
   
{\bf return} Unique($\bigcup_{i=1}^{k}{\mathit{Sols}_i}$) \label{alg:high-level:merge}
\end{algorithm}

\textbf{Our approach:}
To address the above-mentioned shortcoming, this research designs a novel upper-bounding mechanism that effectively reduces the search effort in each parallelized \textsf{MOA*} by establishing a unique information-sharing pipeline between the threads.
Let ($f_1,f_2,\dots,f_k$) be the ${\bf cost}$ of the solution node $x$ obtained in the first thread guided by $\mathit{cost}_1$.
The search in this thread prunes paths with estimated ($\mathit{cost}_2,\dots,\mathit{cost}_k$) no better than ($f_2,\dots,f_k$) due to the first dimension already being non-decreasing.
Now, assume that the second thread, guided by $\mathit{cost}_2$, has just extracted a node $y$ with the primary cost of $f'_2$.
We claim that the dimension in the dominance pruning with $x$ (in the first thread) can be further reduced if we observe $f_2 \leq f'_2$, that is, as soon as we see $f_2$ within the explored range of the second thread.
In this case, we just need to check the estimated ($\mathit{cost}_3,\dots,\mathit{cost}_k$) of paths in the first thread against ($f_3,\dots,f_k$) with the second dimension also removed.
This pruning is correct, as it basically means extension of paths with estimated costs no smaller than ($f_3,\dots,f_k$) -- in any dimension -- would definitely lead to a $\mathit{start}$-$\mathit{goal}$ path no better than either $x$ or one of the optimal solutions obtained in the second thread.
Note that when \textsf{MOA*} in the second thread extracts $y$, it guarantees that all solutions with $\mathit{cost}_2$ smaller than $f'_2$ are already captured.

\noindent
\textit{Constant-time upper bounding:}
The truncation strategy above can be applied to the other threads to potentially reduce the dimension in this upper-bound pruning to \textit{one}, enabling a fast \(O(1)\) time pruning rule against a subset of solutions.
For example, if we observe the $f_3$-value of the doubly truncated cost vector ($f_3,\dots,f_k$) within the explored range of the third thread (guided by $\mathit{cost}_3$), with a similar reasoning, we can perform one more truncation and use the vector ($f_4,\dots,f_k$) as a tighter upper bound instead.
It then becomes clear that $k-1$ consecutive truncations of the cost vector ($f_1,\dots,f_k$) yields $f_k$ as a scalar upper bound on the $k$-th objective of the problem. 
This implies that the dimensionality of the problem in upper-bound pruning can potentially be reduced to one, as nodes only need to be checked against a scalar upper bound if a solution node's cost vector has already undergone $k-1$ consecutive truncations.

\noindent
\textit{Global upper bound and early termination:}
Cost vectors truncated $k-1$ times can serve as a \textit{global} upper bound, and also a termination criterion in parallelized \textsf{MOA*}. 
Here, `global' indicates that each scalar upper bound is not restricted to the thread in which it was obtained but can be utilized across the entire search.
Let $\overline{\bf f} = (\overline{f_1},\dots,\overline{f_k})$ represent a vector of scalar upper bounds.
In case of objective ordering with cyclic permutations of costs, $\overline{\bf f}$ can be fully leveraged, as each thread is able to capture one of the scalar upper bounds.
Thus, we can prune nodes (in any of the parallel searches) not respecting at least one of the established global upper bounds. 
Initially set to infinity, these global upper bounds can not only help other threads to strengthen their upper bound pruning, but enable them to terminate early as soon as their primary cost exceeds the corresponding global upper bound.
For instance, the global upper bound $\overline{f_k}$ can be used in the $k$-th thread to terminate the search early as soon as a node with a primary cost no smaller than $\overline{f_k}$ is extracted, given that nodes in the $k$-th thread are always explored in monotonically non-decreasing order of their $f_k$-value.

\begin{algorithm}[!t]

\small
\caption{Parallelized \textsf{MOA*} Search}
\label{alg:pmoa}
\DontPrintSemicolon
 \KwIn{Problem ($\mathit{G}$, $\mathit{\bf cost}$, ${\bf h}$, $\mathit{\mathit{start}}$, $\mathit{\mathit{goal}}$) and a set $\mathit{Sols}$}
 
 \KwOutput{$\mathit{Sols}$ updated with a subset of optimal solutions}
 
$\mathit{Open} \gets \emptyset $\;
$\mathbf{G_{}^{Tr}}(\mathit{u}) \gets \emptyset^{k-1}$ \ $\forall u \in S$ \label{alg:pmoa:init}\;
 $x \gets $ new node with $s(x) = \mathit{start}$\ \;
 $ {\bf g}(x) \gets {\bf 0}$ ,
 $ {\bf f}(x) \gets {\bf h}(\mathit{start})$ ,
 $parent(x) \gets null $ \;
Add $x$ to $Open$\;

\While{$Open \neq \emptyset$ \label{alg:pmoa:iter}}
{
 
Extract node $x$ from $Open$ with the smallest $\textbf{f}$-value \label{alg:pmoa:least_cost} \;

       \If{$\mathtt{IsDominated}(\mathrm{Tr}({\bf g}(x)),\mathrm{G_{1}^{Tr}}(s(x)))$  \textnormal{\bf or} $\mathtt{IsDominated_{MD}}({\bf f}(x),\mathbf{G^{Tr}}(\mathit{goal}))$ \label{alg:pmoa:dom2}} 
      {\textbf{continue} }
      
     $\mathtt{RemoveDominated}(\mathrm{Tr}({\bf g}(x)), \mathrm{G_{1}^{Tr}}(s(x)))$ \;

    Add $\mathrm{Tr}({\bf g}(x))$ to $\mathrm{G_{1}^{Tr}}(s(x))$ \;
     
\If{$ s(x) = \mathit{goal} $ \label{alg:pmoa:goal} \label{alg:pmoa:sol0}}
    {      
        $\mathtt{UpdateUpperBound}(\mathbf{G^{Tr}}(\mathit{goal}))$\label{alg:pmoa:sol2}\;
        Add $x$ to $\mathit{Sols}$ \label{alg:pmoa:sol3}\;
        {\bf continue} \;
    }
    
    \ForEach{$t \in Succ(s(x))$}
        {  $y \gets $ new node with $s(y) = t$ \label{alg:pmoa:expansion1}\; 
             ${\bf g}(y) \gets {\bf g}(x) + {\mathit{\bf cost}} (s(x),t)$ \; 
             ${\bf f}(y) \gets {\bf g}(y) + {\bf h} (t)$ \;
             $parent(y) \gets x$ \label{alg:pmoa:expansion2}\; 

        \If{$\mathtt{IsDominated}(\mathrm{Tr}({\bf g}(y)),\mathrm{G_{1}^{Tr}}(t))$  \textnormal{\bf or} $\mathtt{IsDominated_{MD}}({\bf f}(y),\mathbf{G^{Tr}}(\mathit{goal}))$ \label{alg:pmoa:dom3}} 
      {\textbf{continue} }
            
            Add $y$ to $Open$\; \label{alg:pmoa:expansion3}
            
        }

}
\Return{}
\end{algorithm}
\begin{algorithm}[!ht]
\small
\caption{$\mathtt{IsDominated_{MD}}$}
\label{alg:Isdominated}
\DontPrintSemicolon
\KwIn{A vector $\mathbf{v}$ and a list of sets $\mathbf{V}$}
 
 \KwOutput{$\mathit{true}$ if $\mathbf{v}$ is weakly dominated, $\mathit{false}$ otherwise}

    \For{$i \in \{1, \dots, k\}$}
        {
    \lIf {$f_i(\mathbf{v}) \geq \overline{f_i}$ \label{alg:Isdominated:global}} 
        {\Return{$\mathit{true}$}}
    }
    \For{$\lambda \in \{1, \dots, k-1\}$}
        {  
        \lIf{$\mathtt{IsDominated}(\mathrm{Tr}^\lambda(\mathbf{v}), \mathrm{V}_{\lambda})$}
        {\Return{$\mathit{true}$}}        
        }
\Return{$\mathit{false}$}
\end{algorithm}
\begin{algorithm}[!ht]
\small
\caption{$\mathtt{UpdateUpperBound}$}
\label{alg:updateUB}
\DontPrintSemicolon
\KwIn{List of sets $\mathbf{V}$}
 
 \KwOutput{List of sets $\mathbf{V}$ updated}
 
\For{$\lambda \in \{1,\dots,k-2 \}$}
{
$i \gets$ index of the primary cost of vectors in $\mathrm{V}_{\lambda}$ \;
\lIf{$\mathit{Sols}_i = \emptyset$}{{\bf continue}}
$z \gets$ The most recent solution node in $\mathit{Sols}_i$ \;

\For{$\mathbf{v} \in \mathrm{V}_{\lambda}$}
{
    \If{$f_i(\mathbf{v} ) \leq f_i(z)$}
    {
    Remove $\mathbf{v}$ from $\mathrm{V}_{\lambda}$ \;
    \If{$\mathtt{IsDominated}(\mathrm{Tr}(\mathbf{v}), \mathrm{V}_{\lambda+1})$\label{alg:updateUB:dom}}
    {{\bf continue}}
    
    $\mathtt{RemoveDominated}(\mathrm{Tr}(\mathbf{v}),\mathrm{V}_{\lambda+1})$\label{alg:updateUB:rem}\;

    Add $\mathrm{Tr}(\mathbf{v})$ to $\mathrm{V}_{\lambda+1}$\label{alg:updateUB:add}\;
    }
}
}
\If { $\mathrm{V}_{k-1} \neq \emptyset$\label{alg:updateUB:global1}}
{
$i \gets$ index of the scalar cost in $\mathrm{V}_{k-1}$ \;
$\overline{f_i} \gets$ the scalar cost in $\mathrm{V}_{k-1}$\label{alg:updateUB:global2}\;
}
\Return{}
\end{algorithm}

To implement our upper bounding strategy, we initialize a list of solution sets $\mathit{\bf Sols}$ as $\{\mathit{Sols}_1,\dots,\mathit{Sols}_k\}$, and a global upper bound vector $\overline{\bf f}$, both shared between all threads (line~\ref{alg:high-level:init} of Algorithm~\ref{alg:high-level}).
We now explain the necessary changes in \textsf{MOA*} to incorporate the upper bounding procedure.
Algorithm~\ref{alg:pmoa} shows the detailed steps involved in our $i$-th parallelized \textsf{MOA*} led by $\mathit{cost}_i$. 
There are three key changes:\\
i) To facilitate multidimensional truncations, $\mathbf{G_{}^{Tr}}$ now organizes truncated cost vectors by storing those with the same cardinality together in separate lists.
Thus, we have $\mathbf{G_{}^{Tr}}$ as $\{\mathrm{G_{1}^{Tr}},\dots,\mathrm{G_{k-1}^{Tr}}\}$ with the index of $\mathrm{G_{}^{Tr}}$ denoting the number of truncations.
For example, $\mathrm{G_{2}^{Tr}}$ contains (non-dominated) doubly truncated costs vectors.
Although the multidimensional truncation is applied to the cost vectors of the solution nodes (associated with $\mathit{goal}$) only, we generalize it to all states for the sake of simplicity in our algorithm description (line \ref{alg:pmoa:init}).\\
ii) 
Dominance check against previous solutions (our upper bounds) is now done through the $\mathtt{IsDominated_{MD}}$ procedure
(lines~\ref{alg:pmoa:dom2}, \ref{alg:pmoa:dom3}).
As presented in Algorithm~\ref{alg:Isdominated}, the cost vector $\mathbf{v}$ is first checked against the global upper bound $\overline{\bf f}$, where $f_i(\mathbf{v})$ denotes the $f_i$-value of the cost vector.
There will then be $k-1$ phases of dominance checks if $\mathbf{v}$ is deemed within the global upper bound.
In each phase, the $\lambda$ truncated cost vector is checked against a corresponding set of truncated costs in $\mathbf{G_{}^{Tr}}$, namely $\mathrm{G_{\lambda}^{Tr}}(\mathit{goal})$.
\\
iii) Upon extraction of a non-dominated solution node, the search attempts to improve the upper bounds stored in $\mathbf{G_{}^{Tr}}(\mathit{goal})$ through the $\mathtt{UpdateUpperBound}$ procedure (line~\ref{alg:pmoa:sol2}), depending on the search progress in the other threads.
Central to our upper bounding strategy, this procedure is detailed in Algorithm~\ref{alg:updateUB}.
In each thread, starting from the very first $\mathrm{G_{1}^{Tr}}(\mathit{goal})$ set, the procedure checks vectors of each set indexed by $\lambda \in \{1,\dots,k-2\}$ for a further truncation.
Let $i$ be the primary (first appearing) cost index of the vector $\mathbf{v}$ in $\mathrm{G_{\lambda}^{Tr}}(\mathit{goal})$. 
The procedure then retrieves the most recent solution of the $i$-th thread from $\mathit{Sols}_i$ (or alternatively, the most recently extracted node in the $i$-th thread). 

The vector $\mathbf{v}$ can be further truncated if we observe that its $f_i$-value is no larger than the $f_i$-value of the retrieved solution (or the recently extracted) node from the $i$-th thread.
The truncation involves: removing the vector from its current set $\mathrm{G_{\lambda}^{Tr}}$, truncating it, and inserting it into the next set $\mathrm{G_{\lambda+1}^{Tr}}$, while ensuring the vectors of the set remain non-dominated after the new vector is added.
Thus, the last set $\mathrm{G_{k-1}^{Tr}}$ is either empty or contains a single scalar.
To keep the vectors of each set non-dominated, we first check the newly truncated vector against the existing candidates in $\mathrm{G_{\lambda+1}^{Tr}}$.
If the new vector is dominated, it does not need to be added to the set (line~\ref{alg:updateUB:dom} of Algorithm~\ref{alg:updateUB}).
Otherwise, the newly truncated vector is non-dominated and should be added to the set.
To keep our future upper-bound mechanism efficient, we remove from $\mathrm{G_{\lambda+1}^{Tr}}$ all its vectors weakly dominated by the new vector (line~\ref{alg:updateUB:rem}).
Compared with the single truncation technique utilized in the recent \textsf{MOA*} approaches, removal of dominated vectors from each set allows our upper-bounding method to reduce the total number of costs vectors stored in $\mathbf{G_{}^{Tr}}(\mathit{goal})$.
The newly truncated vector will be added into $\mathrm{G_{\lambda+1}^{Tr}}$ if it is deemed non-dominated (line~\ref{alg:updateUB:add}).
In the end, if $\mathrm{G_{k-1}^{Tr}}$ is not empty, the procedure utilizes the (single) scalar upper bound obtained through $k-1$ truncations to update the corresponding global upper bound (lines~\ref{alg:updateUB:global1}-\ref{alg:updateUB:global2}).

The following example further elaborates on our proposed multidimensional truncation technique and demonstrates its effectiveness in reducing the search space.
\begin{table*}[htbp]
\centering
    \begin{tabular}{c c c c}
        \begin{tabular}{|c|}
            \toprule
             $\mathit{Sols_1}$\\ \midrule
            $(f_1,f_2,f_3,f_4)$ \\ \midrule
            $(4, 7, 8, 8)$ \\ 
            $(5, 5, 9, 4)$ \\ 
            $(6, 9, 8, 7)$ \\
            $(7, 8, 5, 8)$ \\
            $(8, 4, 7, 9)$ \\
            \textcolor{red}{$(9, 5, 6, 8)$} \\
            \bottomrule
            \multicolumn{1}{c}{(a)}
        \end{tabular}
        $\Rightarrow$
        \begin{tabular}{|c|c|c|}
            \toprule
            $\mathrm{G_1^{Tr}}$& $\mathrm{G_2^{Tr}}$& $\mathrm{G_3^{Tr}}$ \\\midrule
            $(f_2,f_3,f_4)$ & $(f_3,f_4)$ & $f_4$ \\ \midrule
            \cancel{$(7 ,8 , 8)$} & & \\ 
            $(5, 9, 4)$ & & \\ 
            $(9, 8, 7)$ & & \\
            $(8, 5, 8)$ & & \\
            $(4, 7, 9)$ & & \\
            $(5, 6, 8)$ & & \\
            \bottomrule
            \multicolumn{3}{c}{(b) First truncation phase}
        \end{tabular}
        $\to$
        \begin{tabular}{|c|c|c|}
            \toprule
            $\mathrm{G_1^{Tr}}$& $\mathrm{G_2^{Tr}}$& $\mathrm{G_3^{Tr}}$ \\\midrule
            $(f_2,f_3,f_4)$ & $(f_3,f_4)$ & $f_4$ \\ \midrule
            & & \\
            \cancel{$(5, 9, 4)$} & $(9, 4)$ & \\ 
            $(9, 8, 7)$ &  & \\
            $(8, 5, 8)$ &  & \\
            \cancel{$(4, 7, 9)$} & \cancel{$(7, 9)$} & \\
            \cancel{$(5, 6, 8)$} & $(6, 8)$ & \\
            \bottomrule
            \multicolumn{3}{c}{(c) Second truncation phase}
        \end{tabular}
        $\to$
        \begin{tabular}{|c|c|c|}
            \toprule
            $\mathrm{G_1^{Tr}}$& $\mathrm{G_2^{Tr}}$& $\mathrm{G_3^{Tr}}$ \\\midrule
            $(f_2,f_3,f_4)$ & $(f_3,f_4)$ & $f_4$ \\ \midrule
            & & \\
             & $(9, 4)$ & \\ 
            $(9, 8, 7)$ &  & \\
            $(8, 5, 8)$ &  & \\
            &  & \\
            & \cancel{$(6, 8)$} & $8$\\ 
            \bottomrule
            \multicolumn{3}{c}{(d) Third truncation phase}
        \end{tabular}
        
    \end{tabular}
\caption{(a) A sample scenario with six solutions in the thread guided by $f_1$ (the most recent solution shown in red). Assume $f_2=6$ and $f_3=7$ for the most recent solution of the second and third threads, respectively; (b) The cost vectors in $\mathbf{G_{}^{Tr}}$ after the initial truncation in \textsf{MOA*}. The truncated vector of the recent solution dominates that of the first solution (crossed-out); (c) The second phase further truncates three of the vectors (crossed-out) due to having their $f_2$within the explored range of the second thread; (d) The last phase similarly truncates one of the (doubly truncated) vectors, yielding an upper bound of $8$ on $f_4$. 
}
\label{table:example}
\end{table*}

\textbf{Example:}
Consider a MOSP instance with four objectives ($k=4$) to be solved with our parallel framework.
There will be four threads, each guided by one of the objectives as a primary cost.
The first thread takes the conventional $(\mathit{cost}_1,\dots,\mathit{cost}_4)$ order.
For the search in this thread, assume we have already found six solution paths, all provided in Table~\ref{table:example}(a).
$\mathit{cost}_2$ and $\mathit{cost}_3$ are the primary cost of the second and third search, respectively. 
Assume we have $f_2=6$ for the most recently solution in the second thread, and also $f_3=7$ for the latest solution in the third thread.
We now show how the first thread can use the search progress in the other threads to improve its upper bounds in $\mathbf{G_{}^{Tr}}(\mathit{goal})$ via three consecutive truncation phases.
\\
\textit{First truncation:}
\textsf{MOA*} conventionally stores the non-dominated truncated cost vector of the solutions in a separate list, shown in the left column of Table~\ref{table:example}(b).
We observe that, upon capturing the last solution (shown in red), the first thread removes the very first truncated vector (crossed-out) from the list as we have $(5,6,8) \preceq (7,8,8)$.
Our method now extends the truncation process in additional phases.\\
\textit{Second truncation:}
There are three (singly) truncated vectors in the first column of Table~\ref{table:example}(b) with their $f_2$-value no larger than the $f_2$-value of the most recent solution in the second thread, i.e., $f_2\leq6$.
These vectors are truncated and consequently moved into a separate list containing doubly truncated vectors containing $(f_3,f_4)$, as shown in Table~\ref{table:example}(c).
Once transferred, we realize that the second vector is dominated by the third one, as we have $(6,8) \preceq (7,9)$. 
Thus, the dominated vector is removed to keep the list non-dominated.\\
\textit{Third truncation:}
The middle column of Table~\ref{table:example}(c) contains two doubly truncated vectors, one of which has an $f_3$-value no larger than the $f_3$-value of the retrieved solution of the third thread, i.e., $f_3\leq7$.
This vector will be truncated and consequently moved to the list of triply truncated vectors, as shown in Table~\ref{table:example}(d), yielding an upper bound of $8$ on $\mathit{cost}_4$.

\textit{Global upper bound:}
Since the $\mathtt{UpdateUpperBound}$ procedure tracks scalar upper bounds obtained in each parallel search, the single scalar cost from the final truncation phase in Table~\ref{table:example}(d) can now serve as a global upper bound on $\mathit{cost}_4$.
Thus, we can have $\overline{f_4} \gets 8$, meaning that nodes with $f_4$-value no smaller than $\overline{f_4}$ can be pruned in all threads.
This upper bound can also serve as a termination criterion for the thread guided by $\mathit{cost}_4$, allowing the thread to halt once it extracts a node with an $f_4$-value of $8$ or greater. 
Note that $\overline{f_4}$ may later be updated (reduced) as additional solution costs are truncated throughout the search.

\textit{Upper-bound pruning:}
With the multidimensional list $\mathbf{G_{}^{Tr}}(\mathit{goal})$ updated, now assume the first thread has extracted a new node with ${\bf f}=(9,4,9,5)$ in the next iteration.
We can observe that this node would not be pruned by any truncated cost vector in Table~\ref{table:example}(a) if we had only applied \textsf{MOA*}'s single truncation phase, potentially leading to finding \textit{duplicate} solutions.
Instead, let us check this new cost vector against the upper bounds obtained in Table~\ref{table:example}(d) through consecutive truncations.
The first truncation yields $(4,9,5)$. None of the two candidates in $\mathrm{G_{1}^{Tr}}(\mathit{goal})$, the left most column of Table~\ref{table:example}(d), dominates the (singly) truncated cost vector.
Thus, we further truncate the vector and obtain $(9,5)$. Checking this doubly truncated vector against the only vector of $\mathrm{G_{2}^{Tr}}(\mathit{goal})$, the middle column of the table, we find it dominated due to having $(9,4) \preceq (9,5)$.
Thus, the new node can be pruned via our upper bounding approach, reducing the expansion effort and search space.

\textbf{Merging optimal solutions:}
Although each parallelized search in our framework is equipped with an upper bounding technique that helps shrink the overlap between the search spaces, it is still possible for the searches to capture some duplicate solutions.
This is because our pruning strategy does not rigorously check nodes against all solutions of other threads.
Thus, upon the parallel loop termination (i.e., once one of the threads terminates), the last step in our framework involves merging the solutions and then returning the cost-unique ones (line~\ref{alg:high-level:merge} of Algorithm~\ref{alg:high-level}).
One straightforward strategy for this task is to first sort all discovered solutions lexicographically by their $\mathit{\bf cost}$, and then perform a linear traversal to remove duplicates.

\section{Theoretical Results}
This section formalizes the correctness of our framework.

\noindent \textbf{Lemma~1\ }
Let $\mathbf{v}$ be the cost vector of a solution node $x$ truncated $\lambda$ times, $\lambda \in \{0,\dots,k-2\}$, in a parallelized \textsf{MOA*}, that is, $\mathbf{v} = \mathrm{Tr}^\lambda({\bf f}(x))$.
Also let $i$ be the index of the primary (first appearing) cost of $\mathbf{v}$, and $f'$ be the $f_i$-value of the most recent node extracted in a thread guided by $\mathit{cost}_i$.
The vector $\mathbf{v}$ can be further truncated if we observe $f_i(x) \leq f'$, which yields pruning nodes whose $\lambda+1$ truncated ${\bf f}$-vector is weakly dominated by $\mathrm{Tr}^{\lambda+1}({\bf f}(x))$.

\noindent \textbf{Proof Sketch\ }
We use mathematical induction. \\
\textit{Base case $\lambda = 0$}:
The very first phase of truncation here is in fact the standard (single) truncation in \textsf{MOA*}. 
$\lambda = 0$ means that the vector $\mathbf{v}$ has not been truncated yet, and the solution node $x$ already belongs to the thread led by $\mathit{cost}_i$.
Given that the $f_i$-value of the extracted nodes in the thread led by $\mathit{cost}_i$ are monotonically non-decreasing, we always have $f_i(x) \leq f'$.
It follows that checking new nodes against the primary cost of $\mathbf{v}$ (for dominance and upper bounding) is not necessary, and thus $\mathbf{v}$ can be truncated.
\\
\textit{Inductive case}:
Assume the pruning rule is correct for
$\lambda$ consecutive truncations.
We now check the correctness of pruning rule for the next truncation phase $\lambda+1$.
Let $\mathbf{v}'$ be the truncated cost vector of $x$ after $\lambda+1$ truncations, i.e., we have $\mathbf{v}'=\mathrm{Tr}^{\lambda + 1}({\bf f}(x))$ and $f_i(x) \leq f'$, where $i$ is the index of the recently truncated cost, or equivalently the index of the primary cost of $\mathbf{v}$.
Assume, for contradiction, that $y$ could lead to a cost-unique optimal solution but is wrongly pruned due to the recent ${\lambda + 1}$ truncation.
$y$ is pruned because we have observed $\mathrm{Tr}^{\lambda+1}({\bf f}(x)) \preceq \mathrm{Tr}^{\lambda+1}({\bf f}(y))$.
We can distinguish two cases:\\
i) $f_i(x) \leq f_i(y)$: 
this immediately yields $\mathrm{Tr}^{\lambda}({\bf f}(x)) \preceq \mathrm{Tr}^{\lambda}({\bf f}(y))$, which means that the weak dominance rule holds and thus $y$ has correctly been pruned based on the assumed pruning rule of the previous truncation phase $\lambda$. \\
ii) $f_i(y) < f_i(x)$:
incorporating the truncation condition, i.e., $f_i(x) \leq f'$, we will have $f_i(y) < f'$.
Given the correctness of \textsf{MOA*} and that the thread guided by $\mathit{cost}_i$ always explores nodes in monotonically non-decreasing order of their $f_i$-value, if $y$ leads to a cost-unique optimal solution, it must have already been captured by the search. 
This is because its $f_i$-value would fall within the explored range of the thread led by $\mathit{cost}_i$, i.e., $f'$.\\
We observe that node $y$ cannot lead to a cost-unique solution in either of the cases above, which contradicts our initial assumption. Thus, we conclude that $\mathbf{v}$ can be safely truncated if $f_i(x) \leq f'$.
 \hfill $\square$

\noindent \textbf{Lemma~2\ }
Let $\mathbf{v}$ be the cost vector of a solution node truncated $\lambda$ times, $\lambda \in \{0,\dots,k-1\}$.
$\mathbf{v}$ is a global upper bound across all threads.

\noindent \textbf{Proof Sketch\ }
Under the premises of Lemma~1, $\mathbf{v}$ can be used within the same thread to prune out-of-bound paths.
However, due to the overlap between search spaces in the parallel setting, it is likely that such paths will also be generated by other threads.
Let $y$ be a node, generated in either of the threads, whose vector of partial costs $\mathbf{v}'$ corresponds with the costs in $\mathbf{v}$.
If $\mathbf{v} \preceq \mathbf{v}'$, with a similar reasoning, we can prune $y$ while guaranteeing that there is at least one solution (in one of the threads) that weakly dominates $y$.
 \hfill $\square$

While the cost vectors of discovered solutions in each thread conventionally serve as upper bounds for their corresponding \textsf{MOA*}, Lemma 2 demonstrates that, in our parallel setting, each individual truncated cost vector can potentially act as a global upper bound, which can be leveraged by any thread to further reduce the search space.
In this study, we only use cost vectors truncated $k-1$ times and form a global upper-bound vector $\overline{\bf f}=(\overline{f_1},\dots,\overline{f_k})$.
It follows that \textsf{MOA*} in the thread led by $\mathit{cost_i}$, $i \in \{1,\dots,k\}$ can terminate early if it extracts a node with an $f_i$-value no smaller than $\overline{f_i}$, knowing that all unexplored nodes in this thread are guaranteed to violate the global upper bound.

\noindent \textbf{Theorem 1\ }
Parallelized \textsf{MOA*} returns a set of cost-unique optimal solution paths to a MOSP instance.

\noindent \textbf{Proof Sketch\ }
Our framework uses $k$ CPU threads to parallelize $k$ individual \textsf{MOA*} on different objective orders, where each thread can solve MOSP optimally.
We proved in Lemmas~1-2 why our proposed upper-bound pruning strategies are correct.
Thus, we just need to show that Algorithm~\ref{alg:high-level} returns a cost-unique optimal solution set when it terminates.
The parallel loop stops as soon as one of the threads terminates.
This termination criterion is correct, as it essentially means the thread has completed a full exploration of the search space.
Each thread returns a subset of cost-unique optimal solutions.
However, due to the overlap between the searches, and since our framework does not fully share upper bounds between the threads, duplicate solutions are likely to appear among the returned solution sets.
Thus, our approach undertakes a separate, yet straightforward sorting task to merge and remove duplicate solutions (line~\ref{alg:high-level:merge} of Algorithm~\ref{alg:high-level}).
Once the cost-unique solutions are extracted, they will be returned as an optimal answer to the given MOSP instance.
 \hfill $\square$

\begin{table}[t]
\centering
\small
\setlength{\tabcolsep}{4.3pt}
\begin{tabular}{|l | r | *{3}{r} | *{1}{r} | *{1}{r} |}
\toprule
      & & \multicolumn{3}{c|}{Runtime(s)} & \multicolumn{1}{c|}{Sp. } & \multicolumn{1}{c|}{Mem. } \\ \cline{3-5}
     Method & $|\mathcal{S}|$ & \multicolumn{1}{c}{Min.} & \multicolumn{1}{c}{Mean} & \multicolumn{1}{c|}{Max.} & \multicolumn{1}{c|}{Up} & \multicolumn{1}{c|}{(GB)}\\

\midrule
\multicolumn{7}{|c|}{NY with 3 cost components (avg($|\mathit{Sols}|$) = 5,090)} \\
\textsf{NWMOA*} & 100  & 0.01          & 2.43 & 14.0 & - & 0.05 \\
\textsf{NWMOA*\textsubscript{par}}   & 100  & 0.01 & 1.70  & 9.2  & 1.31 & 0.11\\
\textsf{LTMOA*} & 100  & 0.03          & 10.60 & 70.0 & - & 0.07  \\
\textsf{LTMOA*\textsubscript{par}}   & 100  & 0.02 & 7.34 & 47.7 & 1.39 & 0.12  \\

\midrule
\multicolumn{7}{|c|}{NY with 4 cost components (avg($|\mathit{Sols}|$) = 86,134)} \\
\textsf{NWMOA*} & 98  & 0.09          & 508.00  & 3600.0 & - & 0.50  \\
\textsf{NWMOA*\textsubscript{par}}   & 100  & 0.07  & 189.06  & 1558.5 & 3.01 & 1.52  \\
\textsf{LTMOA*} & 90  & 0.22          & 988.87 & 3600.0  & - & 0.63 \\
\textsf{LTMOA*\textsubscript{par}}   & 100  & 0.13 & 403.74  & 3282.9 & 2.91 & 1.68  \\

\midrule
\multicolumn{7}{|c|}{NY with 5 cost components (avg($|\mathit{Sols}|$) = 158,898)} \\
\textsf{NWMOA*} & 74  & 0.13          & 1405.49 & 3600.0  & - & 0.52\\
\textsf{NWMOA*\textsubscript{par}}   & 91  & 0.08 & 745.43 & 3600.0 & 4.23 & 2.60\\
\textsf{LTMOA*} & 68  & 0.34          & 1720.05 & 3600.0  & - & 0.66\\
\textsf{LTMOA*\textsubscript{par}}   & 86  & 0.13 & 1034.35 & 3600.0 & 3.75 & 2.89 \\

\bottomrule
\end{tabular}
\caption{Performance of algorithms over 100 instances in three scenarios. Runtime statistics are in seconds and $|\mathcal{S}|$ is the number of solved cases. 
We also report, for mutually solved cases, the average memory usage (in GB) and speed up achieved by parallelization w.r.t. the standard variant.
}
\label{table:results_main}
\end{table}
\section{Experimental Results}
This section evaluates the performance improvement of our proposed parallel framework based on two recent \textsf{MOA*} algorithms: \textsf{NWMOA*}\cite{AhmadiSHJ24} and \textsf{LTMOA*} with lazy dominance tests \cite{LTMOA3}.
We implemented all algorithms in C++ using the same data structure to maintain search nodes, enabling a head-to-head comparison.

\textbf{Benchmark setup:}
For the benchmark map, following the literature, we used the New York map from the 9th DIMACS Implementation Challenge: Shortest Paths\footnote{http://www.diag.uniroma1.it/~challenge9/download.shtml} in scenarios with three to five cost components.
The first and second costs of each edge are \textit{distance} and \textit{time}, respectively.
The third cost is the positive height difference of the endpoints of each link, with height information obtained from Shuttle Radar Topography Mission\footnote{https://www2.jpl.nasa.gov/srtm/}.
For the fourth cost, as in \citet{EMOA22}, we calculate the average (out)degree of the link, that is, the number of adjacent vertices of each end point.
Motivated by hazardous material transportation \cite{erkut2007hazardous}, this cost is designed as an indicator of how interconnected an edge is within the network. The more connections an edge has with other links, the higher the potential risk associated with transporting through the link.
Following \citet{TMDA23}, we assign a value of one to the fifth cost of each edge, with $cost_5$ of paths representing the total number of links (or intersections) traversed.
We then evaluated all algorithms on 100 random ($\mathit{start},\mathit{goal}$) pairs from \citet{AhmadiSHJ24} for three scenarios: $k=3,4,5$.

\textbf{Implementation:} All C++ code was compiled using the GCC7.5 compiler with O3 optimization settings, and all experiments were conducted on an Intel Xeon Platinum 8488C processor with four CPU cores (eight threads with hyper-threading enabled) running at 2.4~GHz and with 16~GB of RAM under Linux environment with a one-hour timeout.
For parallel algorithms, we performed three runs of each algorithm and stored the run showing the median runtime.
For a problem instance with $k$ objectives, we set the parallel algorithms to work on $k$ different orderings obtained by cyclic permutations of objectives.
We then allocated one thread to each objective ordering to fully leverage parallel computation of solutions.
Although the number of threads involved in the computation here is limited to the number of objectives, one can still incorporate available threads for running additional MOA* on a different set of objective orderings.
While this extension allows for higher utilization of resources, the effectiveness of the framework in reducing the overall computation time should be investigated due to added algorithmic overhead.
We will study this extension in future work.
Our code is publicly available\footnote{https://bitbucket.org/s-ahmadi/multiobj}.

\textbf{Performance impact of parallelization:}
Since all algorithms use the same approach to compute heuristic functions, we focus on the performance of the main search only.
Table~\ref{table:results_main} compares the performance of our parallelized algorithms against their standard version, which use lexicographical ordering of objectives.
We report the number of solved cases ($|\mathcal{S}|$ out of 100) and runtime statistics (in seconds).
The runtime of unsolved cases is considered to be the timeout.
We also report, for mutually solved cases, the average speed up (Sp. Up) obtained over the standard variant, and average memory usage (in GB).
We observe that, compared to the standard methods, the parallel variants:\\
i) consistently solve more instances.
In the $k=5$ scenario, \textsf{LTMOA*\textsubscript{par}} and \textsf{NWMOA*\textsubscript{par}} solve 18 and 17 more instances within the one-hour timeout, respectively;\\
ii) offer statistically considerable runtime improvement.
In the $k=4$ scenario, maximum runtime of \textsf{NWMOA*} is reduced from (above) 60 minutes to 26 minutes, and the average runtime from 8 to 3 minutes, approximately;\\
iii) achieve a speed-up proportional to the problem dimension.
Comparing the average speed-up factors across the scenarios, we observe a steady increase as the number of objectives rises, with the average factor growing from 1.3 in the $k=3$ scenario to around 4 in the $k=5$ scenario for both algorithms;\\
iv) require more space to accommodate the scaled search space.
Memory usage increases proportionally as well. 
The parallelized \textsf{LTMOA*} consumes 4.4 times more memory in the $k=5$ scenario, but only 1.7 times in the $k=3$ scenario.


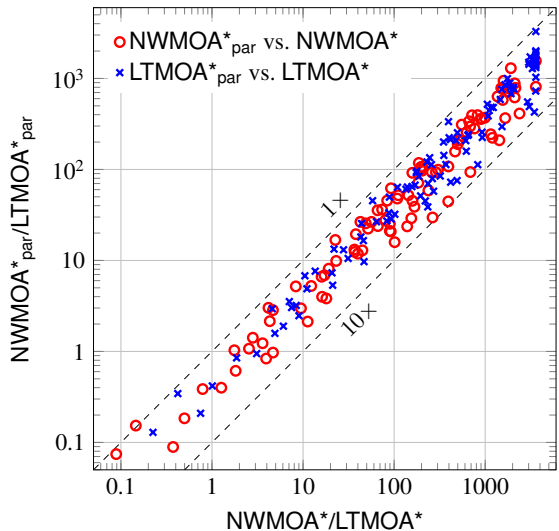
\begin{figure}[t]
    \centering
    \begin{tikzpicture}[scale=0.90]
        \begin{axis}[
            width=\columnwidth,
            height=\columnwidth,
            xlabel={\textsf{NWMOA*}/\textsf{LTMOA*}},
            ylabel={\textsf{NWMOA*\textsubscript{par}}/\textsf{LTMOA*\textsubscript{par}}},
            xmode=log,
            ymode=log,
            xmin=0.05, xmax=6000,
            ymin=0.05, ymax=6000,
            xticklabels={0.01,0.1,1,10,100,1000},
            yticklabels={0.01,0.1,1,10,10\textsuperscript{2},10\textsuperscript{3}},
            log basis x={10},
            log basis y={10},
            legend pos=north west,
            legend cell align=left,
            grid=major,
            axis on top,
            legend style={draw=none},
        ]
        
        \addplot[
            only marks,
            mark=o,
            red,
            mark size=2pt,
            line width=1.0pt,
            fill opacity=0.7,
        ]
        table[x="NWMOA", y="NWMOA_par", col sep=comma] {Figs/scatter_data.csv};
        \addlegendentry{\textsf{NWMOA*\textsubscript{par}} vs. \textsf{NWMOA*}}

        \addplot[
            only marks,
            mark=x,
            blue,
            mark size=2pt,
            line width=1.0pt,
            fill opacity=0.7,
        ]
        table[x="LTMOA", y="LTMOA_par", col sep=comma] {Figs/scatter_data.csv};
        \addlegendentry{\textsf{LTMOA*\textsubscript{par}} vs. \textsf{LTMOA*}}

        \addplot[domain=0.05:6000, samples=100, dashed] {x} 
            node[pos=0.55, above, sloped] {$1\times$};
        
        \addplot[domain=0.05:6000, samples=100, dashed] {0.1*x} 
            node[pos=0.55, below, sloped] {$10\times$};
        
        \end{axis}
    \end{tikzpicture}
    \caption{Runtime distribution of \textsf{NWMOA*} and \textsf{LTMOA*} versus their parallelized variant over instances with $k=4$.}
    \label{fig:scatter_plot}
\end{figure}


%
To better illustrate the runtime improvement achieved by parallelization, we present in Figure~\ref{fig:scatter_plot} the runtime distribution of each parallel algorithm against the base variant with $k=4$.
We observe that, in almost every individual instance, both \textsf{LTMOA*} and \textsf{NWMOA*} have performed faster when parallelized with our framework, reducing the computation time by up to an order of magnitude.

\begin{figure}[!t]
    \includegraphics[width=\columnwidth]{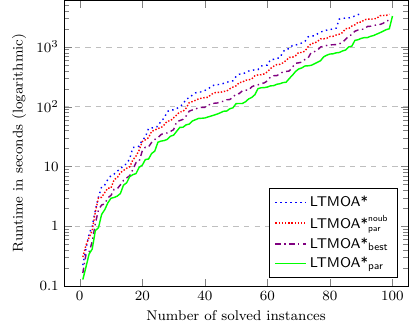}
    \caption{Cactus plot of \textsf{LTMOA*}'s performance versus its virtual best variant, parallelized variant with and without upper bounding (denoted {\textsf{\tiny noub}}).
    Instances are sorted by runtime.
    }
    \label{fig:performance}
\end{figure}

\textbf{Impact of upper-bounding:}
To study the impact of our proposed upper bounding approach on achieving an accelerated search, and also to compare the parallel framework with other variants of the base algorithms, we evaluated \textsf{LTMOA*} over the instances with four objectives, in two other configurations: i) parallelized \textsf{LTMOA*} but with the upper bounding switched off; ii) a virtual best oracle of \textsf{LTMOA*}, which delivers the best runtime among (four) cyclic permutations of objectives. To set up this virtual best oracle, we ran for each instance four runs of the standard \textsf{LTMOA*}, each on a different objective order, and gave the virtual oracle the best runtime.
Figure~\ref{fig:performance} compares the runtime performance of these two variants (denoted by \(\textsf{LTMOA*}_{\textsf{\tiny par}}^{\textsf{\tiny noub}} \) and \textsf{LTMOA*\textsubscript{best}}) against the standard and parallel variants.
There are two key observations that underscore the importance of upper bounding. 
First, the parallel variant outperforms all other variants, including the virtual best variant. Second, although parallelized search with disabled upper bounding can still outperform the standard variant, it falls short of achieving the performance of the virtual best variant. This is primarily due to the considerable search overhead inherent in parallelization.
Thus, we can conclude that the proposed upper bounding technique plays a crucial role in enabling the efficient parallelization of \textsf{MOA*}.

We also evaluated for \textsf{NWMOA*} a (non-parallel) sequential variant of our framework, where the algorithm alternates between the search orders upon discovering a solution in one ordering.
This variant offered only marginal improvements over the standard version, underscoring that the framework is more suited to scenarios where parallelization is feasible.

\section{Conclusion and Future Work}
This research proposed the first parallel search framework for the multi-objective shortest path problem on the basis of A* search (MOA*).
While being applicable to existing \textsf{MOA*} approaches, our framework allows the space of the problem to be searched with different objective orderings simultaneously.
The research also builds a unique information-sharing pipeline between concurrent searches that can reduce the dimensionality of the problem to one in certain cases.
Two of the leading \textsf{MOA*} algorithms were parallelized based on the proposed approach.
The results demonstrate the success of the proposed framework in reducing the computation time of \textsf{MOA*}, achieving an average speed-up proportional to the problem dimension.

\textbf{Future work:}
The number of threads in our proposed parallel framework is determined by the number of objectives.
However, if there are more threads available, the framework can be extended to utilize some of them through incorporating a different set of objective orderings. 
In this scenario, \textsf{MOA*} in some threads are guided by the same primary cost, leading to significant overlap in their search space. 
Future research could explore mechanisms that leverage mutual information between these threads to enhance upper bounding.

\section{Acknowledgements}
This research was supported by the Department of Climate Change, Energy, the Environment and Water under the International Clean Innovation Researcher Networks (ICIRN) program grant number ICIRN000077.
Mahdi Jalili is supported by Australian Research Council through projects DP240100963, DP240100830, LP230100439 and IM240100042.

\bibliography{references}

\end{document}